\documentclass[sigconf]{acmart}
\usepackage{amsmath}

\usepackage[colorinlistoftodos]{todonotes}
\usepackage{balance}
\newcommand{\independent}{\protect\mathpalette{\protect\independenT}{\perp}}
\def\independenT#1#2{\mathrel{\rlap{$#1#2$}\mkern2mu{#1#2}}}
\AtBeginDocument{%
  \providecommand\BibTeX{{%
    \normalfont B\kern-0.5em{\scshape i\kern-0.25em b}\kern-0.8em\TeX}}}
\usepackage{amsfonts,amssymb}
\setcopyright{acmcopyright}

\begin{document}
\fancyhead{}
%%
%% The "title" command has an optional parameter,
%% allowing the author to define a "short title" to be used in page headers.
\title{Matching in Selective and Balanced Representation Space for Treatment Effects Estimation}

%%
%% The "author" command and its associated commands are used to define
%% the authors and their affiliations.
%% Of note is the shared affiliation of the first two authors, and the
%% "authornote" and "authornotemark" commands
%% used to denote shared contribution to the research.
\author{Zhixuan Chu}
\email{zhixuan.chu@uga.edu}
\affiliation{%
  \institution{University of Georgia}
  \city{Athens}
  \state{Georgia}
}

\author{Stephen L. Rathbun}
\email{rathbun@uga.edu}
\affiliation{%
  \institution{University of Georgia}
  \city{Athens}
  \state{Georgia}
}

\author{Sheng Li}
\email{sheng.li@uga.edu}
\affiliation{%
  \institution{University of Georgia}
  \city{Athens}
  \state{Georgia}
}

%%
%% By default, the full list of authors will be used in the page
%% headers. Often, this list is too long, and will overlap
%% other information printed in the page headers. This command allows
%% the author to define a more concise list
%% of authors' names for this purpose.
\renewcommand{\shortauthors}{Chu, et al.}

\begin{abstract}
The dramatically growing availability of observational data is being witnessed in various domains of science and technology, which facilitates the study of causal inference. However, estimating treatment effects from observational data is faced with two major challenges, missing counterfactual outcomes and treatment selection bias. Matching methods are among the most widely used and fundamental approaches to estimating treatment effects, but existing matching methods have poor performance when facing data with high dimensional and complicated variables. We propose a feature selection representation matching (FSRM) method based on deep representation learning and matching, which maps the original covariate space into a selective, nonlinear, and balanced representation space, and then conducts matching in the learned representation space. FSRM adopts deep feature selection to minimize the influence of irrelevant variables for estimating treatment effects and incorporates a regularizer based on the Wasserstein distance to learn balanced representations. We evaluate the performance of our FSRM method on three datasets, and the results demonstrate superiority over the state-of-the-art methods.

\end{abstract}

\maketitle

\section{Introduction}
\label{intro}

Causal inference from observational data has been a critical research topic across many domains including statistics, computer science, education, public policy, economics, and health care. For example, drug developers need to know whether a new medication is beneficial or harmful in post-market surveillance; a government wants to figure out who would benefit from subsidized job training; and economists want to evaluate how a policy affects unemployment rates. Causal inference is defined as the process of estimating causal effects on units after receiving certain treatments. In the above example of drug discovery, patients are units, medication is the treatment, and the causal effect (or treatment effect) is dependent on the recovery status of patients. Although randomized controlled trials (RCT) are usually considered as the golden standard for causal inference, estimating causal effects from observational data has become an appealing research direction owing to the increasing availability of data and the low costs. 

Researchers have developed various frameworks for causal inference, and the most representative ones include the potential outcome framework ~\cite{splawa1990application,rubin1974estimating} and the structural causal model ~\cite{pearl1995causal,pearl2009causality,pearl2014probabilistic}. The potential outcome framework aims to estimate all potential outcomes under different treatments and then calculate the treatment effects. The structural causal model (SCM) combines components of structural equation models, graphical models, and the potential outcomes framework to make the causal inference. In this paper, we focus on the potential outcome framework. 

When estimating treatment effects from observational data, we face two major challenges~\cite{yao2020survey}, i.e., missing counterfactual outcomes and treatment selection bias. Firstly, in real life, we only observe the factual outcome and never all potential outcomes that would potentially have happened had we chosen other different treatment options. In medicine, for example, we only observe the outcome of giving a patient a specific treatment, but we never observe what would have happened if the patient was instead given an alternative treatment. Secondly, unlike randomized controlled experiments, treatments are typically not assigned at random in observational data. In the medical setting, physicians take a set of factors into account, such as the patient's feedback to the treatment, medical history, and patient health condition, when choosing a therapeutic option. Due to this treatment assignment bias, the treated population may differ significantly from the general population. These two major issues make treatment effects estimation very challenging.

A widely used solution is the matching method, where the missing counterfactual outcome of a unit to a treatment is estimated by the factual outcome of its most similar neighbors that have received that treatment. The dataset including matched samples mimics a randomized controlled trial where the distribution of covariates will be similar between treatment and control groups. The only expected difference between the treatment and control groups is the outcome variable being studied. Compared to regression-based methods such as counterfactual regression ~\cite{shalit2017estimating} and Bayesian additive regression trees~\cite{chipman2010bart}, matching approaches are more interpretable and less sensitive to model specification ~\cite{imbens2015causal}.

Most of existing matching methods are performed in the original covariate space (e.g., Nearest Neighbor Matching~\cite{rubin1973matching}, Coarsened Exact Matching~\cite{iacus2012causal}) or in the one-dimensional propensity score space (e.g., Propensity Score Matching~\cite{rosenbaum1983central}). Although rich information is retained in the original covariate space, it will face the curse of dimensionality and introduce more bias when controlling for irrelevant variables. Theoretical studies revealed that the bias of matching methods increases with the dimensionality of the covariate space~\cite{abadie2006LSP_matching}. Propensity score matching combats the curse of dimensionality of matching directly on the original covariates by matching on the probability of a unit being assigned to a particular treatment given a set of observed covariates. However, a one-dimensional propensity score space will lose most of the information in the data. In addition, provided that models are not over-specified, nonlinear models are usually more capable of dealing with complicated data distributions.

Therefore, learning low-dimensional balanced and nonlinear representations instead of high-dimensional original covariates space or one-dimensional propensity score space for observational data is a promising solution, which has been discussed in~\cite{li2017matching_nips17,chang2017informative_subspace_aaai17}. The major drawback of existing causal inference methods including matching methods is that they always treat all observed variables as pre-treatment variables, which are not affected by treatment assignments but may be predictive of outcomes. This assumption is not tenable for observational data such as post-market pharmaceutical surveillance, cross-sectional studies, electronic medical records, and so on. If all observed variables are directly used to estimate treatment effects, more bias may be introduced into the model. For example, conditioning on an instrumental variable, which is associated with the treatment assignment but not with the outcome except through exposure, can increase both bias and variance of estimated treatment effects ~\cite{myers2011effects_instrumental}. Therefore, conditioning on these variables, let alone irrelevant variables, will introduce more impalpable bias into the model, especially in scenarios with high dimensional variables.

To address the above issues, we propose a deep feature selection representation matching (FSRM) model for treatment effects estimation in representation space. The key idea of FSRM is to map the original covariate space into the selective, nonlinear, and balanced representation space, which is predictive of treatment outcome and treatment assignment, simultaneously. In this way, FSRM could mitigate selection bias and minimize the influence of irrelevant variables by simultaneously predicting the treatment assignment and outcomes. FSRM contains deep feature selection, balanced representation learning, and deep prediction network. Deep feature selection uses a sparse one-to-one layer and deep structures to model non-linearity, selecting a subset of features from the input observational data. The deep prediction network helps learn latent representations that are predictive of treatments and observed outcomes. Due to the treatment assignment bias, there is an imbalance between the original treatment and control distributions. Matching is not always perfect due to incomplete overlap between treatment and control groups, which will lead to biased estimates of treatment effects. To address this issue, the balanced representation learning component of FSRM attempts to reduce the discrepancy between the two distributions by incorporating a regularizer based on the Wasserstein distance ~\cite{sriperumbudur2012empirical}. Finally, FSRM performs matching in the balanced representation space to estimate treatment effects. To the best of our knowledge, FSRM is the first matching method that seamlessly integrates deep feature selection and deep representation learning for causal inference together with the joint prediction of treatment assignment and counterfactual outcomes for causal inference. We evaluate the proposed FSRM method on IHDP, modified IHDP, and simulated datasets, and demonstrate its superiority over the state-of-the-art methods for treatment effects estimation, especially when the data includes different types of variables.

We organize the rest of our paper as follows. Technical background including the basic notations, definitions, and assumptions are introduced in Section 2. Our proposed framework is presented in Section 3. In Section 4, experiments on IHDP, modified IHDP, and simulation datasets are provided. Section 5 reviews related work.

\section{Background}

Suppose that the observational data contain $n$ units and that each unit received one of two or more treatments. Let $t_i$ denote the treatment assignment for unit $i$; $i=1,...,n$. For binary treatments, $t_i = 1$ for the treatment group, and $t_i=0$ for the control group. The outcome for unit $i$ is denoted by $Y_{t}^i$ when treatment $t$ is applied to unit $i$; that is, $Y_{1}^i$ is the potential outcome of unit $i$ in the treatment group and $Y_0^i$ is the potential outcome of unit $i$ in the control group. For observational data, only one of the potential outcomes is observed according to the actual treatment assignment of unit $i$. The observed outcome is called the factual outcome and remaining unobserved potential outcomes are called counterfactual outcomes. Let $X \in \mathbb{R}^d$ denote all observed variables of a unit. 

In this paper, we follow the potential outcome framework for estimating treatment effects ~\cite{splawa1990application, rubin1974estimating}. The individual treatment effect (ITE) for unit $i$ is the difference between the potential treated and control outcomes, and is defined as:
\begin{equation}
    \text{ITE}_i = Y_1^i - Y_0^i, \quad (i=1,...,n).
\end{equation}
The average treatment effect (ATE) is the difference between the mean potential treated and control outcomes, which is defined as:
\begin{equation}
 \text{ATE}=\frac{1}{n}\sum_{i=1}^{n}(Y_1^i - Y_0^i), \quad (i=1,...,n).
\end{equation}

\begin{figure*}[t]
    \centering
    \includegraphics[width=0.8\textwidth]{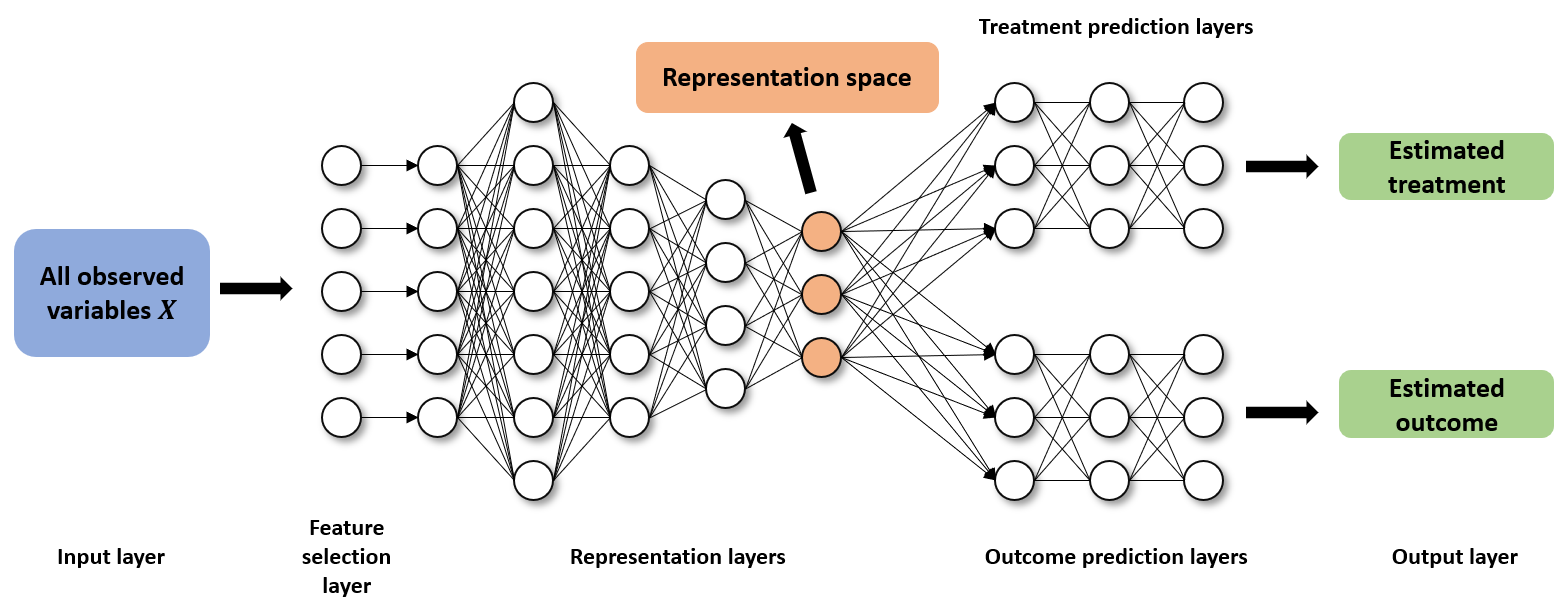}
    \caption{The framework of the proposed feature selection representation matching.}
    \label{fig: framework}
\end{figure*}

The success of the potential outcome framework is based on the following assumptions~\cite{imbens2015causal}, which ensure that the treatment effect can be identified.

\textbf{Stable Unit Treatment Value Assumption (SUTVA)}: The potential outcomes for any units do not vary with the treatments assigned to other units, and, for each unit, there are no different forms or versions of each treatment level, which lead to different potential outcomes. \textbf{Consistency}: The potential outcome of treatment $T$ is equal to the observed outcome if the actual treatment received is $T$. \textbf{Positivity}: For any value of $\,X$, treatment assignment is not deterministic, i.e.,$P(T = t | X = x) > 0$, for all $t$ and $x$. \textbf{Ignorability}: Given covariates $X$, treatment assignment $T$ is independent to the potential outcomes, i.e., $(Y_1, Y_0) \independent T | X$.

\section{The Proposed Framework}

\subsection{Motivation}
\label{Motivation}

Estimating treatment effects from observational data is faced with two major challenges, i.e., missing counterfactual outcomes and treatment selection bias. To overcome these challenges, we aim to propose a new matching method with the following characteristics:

\textbf{Selective.} If all observed variables are directly used to estimate treatment effects, more bias will be introduced into the model as discussed in Section \ref{intro}. To address this issue, it is critical to select important features from observational data. Such a feature selection mechanism will also make the deep neural networks and matching estimators more interpretable.

\textbf{Nonlinear.} Observational data always involve a large number of variables with complicated relationships among them. Under deep neural networks, the data inform the relationship between outcomes and predictors with enough flexibility to describe complicated data distributions. Therefore, deep neural networks could be used to learn nonlinear representations for treated and control units, and then benefit the matching procedure.

\textbf{Adjustable.} The dimension of the representation vector is adjustable according to the complexity of data. It can avoid the curse of dimensionality caused by matching in the high-dimensional original covariate space or information loss caused by one-dimensional propensity scores.

\textbf{Balanced.} Like propensity score matching by controlling for the covariates that predict treatment assignments, matching in the representation space that is predictive of treatment assignments can also reduce the selection bias due to confounding variables. In addition, matching will lead to biased estimates of treatment effects due to incomplete overlap between treatment and control groups. We incorporate Wasserstein distance to measure the distance between representation distributions of treatment and control groups to ensure that the Positivity assumption holds.

\textbf{Individualized.} Most matching methods can only perform well for estimating the average treatment effect but have poor performance in individual treatment effect estimation compared with regression-based methods. The main reason is that matching methods mainly focus on reducing the bias due to confounding variables but neglect the prediction of observed outcomes based on pre-treatment variables. Our representation space can, by predicting observed outcomes, best represent pre-treatment variables, which means that our matching method has competitive performance with respect to individual treatment effect estimation.

\subsection{Model Architecture}
We propose a feature selection representation matching (FSRM) method based on deep representation learning and matching in the representation space. The key idea of FSRM is to map the original covariate space into a selective, nonlinear, and balanced representation space, which can be best predictive of individual treatment outcomes, mitigate selection bias, and minimize the influence of irrelevant variables by simultaneously predicting the treatment assignment and outcomes. The framework of FSRM is illustrated in Fig.~\ref{fig: framework}, which contains five major components: feature selection layer, deep representation layers, outcome prediction layers, treatment prediction layers, and matching in the representation space.

The proposed FSRM method uses the learned representation vectors as the balancing scores to perform matching. The first stage of FSRM adopts a feature selection layer, which is expressed as a nonlinear mapping $\Phi : X \rightarrow R$, where $X$ denotes the original covariate space and $R$ denotes the representation space. In the second stage, FSRM not only predicts the treatment assignment through the function $h_1$, which maps $R$ onto the treatment space  $T=\{0,1\}$, but also predicts the observed outcome through the function $h_2$, which maps $R \times T$ onto the outcome space $Y$.

Moreover, to satisfy the requirement of the matching method that treatment and control groups overlap with respect to covariates, we need to balance the representation distributions between the two groups. The Integral Probability Metric (IPM) ~\cite{shalit2017estimating, sriperumbudur2012empirical} is incorporated into FSRM to maximize the overlap of representation distributions of treatment and control groups. Finally, optimal matching is performed on the selective, nonlinear, and balanced representations. 

Let $\Phi(x)$, $h_1(\Phi(x))$ and $h_2(\Phi(x), t)$ be parameterized by deep neural networks trained jointly in an end-to-end fashion. Deep neural networks are models structured by multiple hidden layers with non-linear activation functions. They can often dramatically increase prediction accuracy, describe complex relationships, and generate the structured high-level representations of features which can assist interpretation of data. In the following, we introduce each component of FSRM in detail.

\subsubsection{Deep Feature Selection}

For  $ \Phi : X \rightarrow R $, we adopt a deep feature selection model ~\cite{li2016deep} that enables variable selection in deep neural networks. This model takes advantage of deep structures to capture data non-linearity and conveniently selects a subset of features of the data at the input level and following representation layers. In this model, the first feature selection layer is a sparse one-to-one layer between the input and the first hidden layer. Feature selection at the input level can help select which variables are input into the neural network and used for representing pre-treatment variables, which makes the deep neutral network more interpretable. 

In the first feature selection layer, every input variable only connects to its corresponding node where the input variable is weighted. This is a 1-1 layer instead of a fully connected layer. To select input features, weights $w$ in the feature selection layer and the following representation layers have to be sparse and only the features with nonzero weights are selected to enter the following layers.

LASSO ~\cite{tibshirani1996regression} was considered first for this purpose. It is a penalized least squares method imposing the $L_1$-penalty on the regression coefficients by $\Re(w) = \lVert w \rVert_1$. However, for observational data with high dimensional variables, LASSO cannot remove enough variables before it saturates. To overcome this limitation, the elastic net ~\cite{zou2005regularization} is adopted in our model, which adds a quadratic term $\lVert w \rVert_2^2$ to the penalty, that is $\Re(w) = \lambda\lVert w \rVert_2^2+\alpha\lVert w \rVert_1$, where $\lambda$ and $\alpha$ are trade-off parameters. Therefore, for feature selection in the deep neural networks of FSRM, we minimize the objective function: 

\begin{equation}
\begin{split}
    \Re = \lambda \sum_{s=1}^{S}\lVert w^{(s)} \rVert_2^2+\alpha\sum_{s=1}^{S}\lVert w^{(s)} \rVert_1,\\
    \label{Eqn: observed and potential outcome}
\end{split}
\end{equation}

where $S$ is the number of hidden layers including the feature selection layer and the representation layers in FSRM. The terms $\lambda \geq 0$ and $ \alpha \geq 0$ are hyper-parameters that not only control the trade-off between regularization term and the following objective terms but also control the trade-off between smoothness and sparsity of the weights in the feature selection layer ~\cite{li2016deep}.

As discussed in this section, we combine two ideas: a sparse one-to-one feature selection layer between the input and the first hidden layer selects which variables are input into the neural network and elastic net throughout the fully-connected representation layers assigns larger weights to important features. This strategy can effectively filter out the irrelevant variables and highlight the important variables.

\subsubsection{Deep Prediction Network}

For $h_1(\Phi(x))$ and $h_2(\Phi(x), t)$, we adopt two branches of deep neural networks to predict the outcomes $Y_i$ and treatment assignments $T_i$ based on the representations $\Phi(x)$, as illustrated in Fig.~\ref{fig: framework}. Each branch is implemented by fully connected layers and one output regression layer. 

The function $h_1(\Phi(x))$ maps the representation vector to the corresponding observed treatment assignment $T_i$. We use the cross entropy loss $\mathcal{L}_T$ to quantify the factual treatment prediction error:
\begin{equation}
\mathcal{L}_T = -\frac{1}{n}\sum_{i=1}^{n}\sum_{j=1}^{K}(t_{ij}log(\hat{t}_{ij})),
\label{Eqn: entropy}
\end{equation}
where $i$ indexes units and $j$ indexes the treatment assignment classes. The terms $t_{ij}$ are the factual probability distributions over $K$ classes for unit $i$ for each treatment assignment $j$, and $\hat{t}_{ij}$ are the predicted probability distributions over $K$ classes for unit $i$. A squared $L_2$ norm regularization term, $\beta \sum_{p=1}^{P}\lVert w^{(p)} \rVert_2^2$, is placed on the model parameters $w^{(p)}$  to mitigate over-fitting, where $\beta \geq 0$ denotes a tuning constant controlling the trade-off between the $L_2$ regularization term of prediction layers and other terms in final objective function.

The function $h_2(\Phi(x), t)$ maps the representation vector and treatment assignment to the corresponding observed outcome. However, when the dimensionality of representation space is high, there is a risk of losing the influence of $t$ on $h_2(\Phi(x), t)$ if the concatenation of $\Phi(x)$ and $t$ is treated as input~\cite{shalit2017estimating}. To solve this issue, $h_2(\Phi(x), t)$ is partitioned into two head layers $h_2^0(\Phi)$ and $h_2^1(\Phi)$:
\begin{equation}
h_2(\Phi(x), t) =  
  \begin{cases} 
   h_2^0(\Phi) & \text{if } t = 0 \\
   h_2^1(\Phi)      & \text{if }  t = 1. 
  \end{cases}
  \label{Eqn: hypothesis}
\end{equation}
Here, the first layer $h_2^1(\Phi)$ is used to estimate the outcome under treatment and the second layer $h_2^0(\Phi)$ is used to estimate the outcome for the control group. Each sample is only updated in the head layer corresponding to the observed treatment. Obviously, this model can also be extended to any number of treatments. Let $\hat{y}_i=h_2(\Phi(x), t)$ denote the inferred observed outcome of unit $i$ corresponding to factual treatment $t_i$. We aim to minimize the mean squared error in predicting factual outcomes
\begin{equation}
 \mathcal{L}_Y = \delta \frac{1}{n}\sum_{i=1}^{N}(\hat{y}_i-y_i)^2, \\
 \label{Eqn: outcome}
\end{equation}
where the hyper-parameter $\delta$ controls the trade-off between the outcome prediction and treatment prediction loss functions. A squared $L_2$ norm regularization term on the model parameters, $\beta \sum_{p=1}^{P}\lVert w^{(p)} \rVert_2^2$, is added to mitigate the overfitting problem, where $\beta \geq 0$ denotes the hyper-parameter controlling the trade-off between the $L_2$ regularization term of the prediction layers and other terms in the final objective function.

\subsubsection{Learning Balanced Representations}

The deep feature selection and deep prediction networks learn compact and nonlinear representations for control and treated units. However, the distributions of the treatment group and control group might be imbalanced in the representation space.  
%we could generate nonlinear representation space for control and treated units. 
For the matching procedure, the representation distributions of treatment and control groups should have overlap. To this end, we adopt integral probability metrics (IPM) when learning the representation space to make sure that the nonlinear representation distributions are balanced for the two groups. The integral probability metrics measure the divergence between the representation distributions of treatment and control groups, so we want to minimize the IPM to make two distributions more similar. Let $P(\Phi(x)|t=1)$ and $Q(\Phi(x)|t=0)$ denote the empirical distributions of the representation vectors for the treatment and control groups, respectively. In FSRM, we adopt the IPM defined in the family of 1-Lipschitz functions, which leads to IPM being the Wasserstein distance ~\cite{shalit2017estimating, sriperumbudur2012empirical,guo2019learning}. In particular, the IPM term with Wasserstein distance is defined as
%(Eq. (~\ref{Eqn: wass})) denotes the,  The IPM term is defined as follows:
\begin{equation}
 \text{Wass}(P,Q) = \gamma \inf_{k \in \mathcal{K}} \int_{\Phi(x)} \lVert k(\Phi(x))-\Phi(x) \rVert P(\Phi(x))d(\Phi(x)),
 \label{Eqn: wass}
\end{equation}
where $\gamma$ denotes the hyper-parameter controlling the trade-off between $\text{Wass}(P,Q)$ and other terms in the final objective function. $\mathcal{K}=\{k|Q(k(\Phi(x))) = P(\Phi(x)) \}$ defines the set of push-forward functions that transform the representation distribution of the treatment distribution $P$ to that of the control $Q$ and $\Phi(x) \in \{\Phi(x)_i\}_{i:t_i=1}$.

\subsubsection{Objective Function}

Putting all the above together, the objective function of our feature selection representation matching (FSRM) model is:

\begin{equation}
 \begin{split}
    \mathcal{L}=
    &-\frac{1}{n}\sum_{i=1}^{n}\sum_{j=1}^{K}(t_{ij}log(\hat{t}_{ij}))\\
    &+\delta \frac{1}{n}\sum_{i=1}^{N}(\hat{y}_i^{t_i}-y_i)^2\\
    &+ Wass(P,Q)\\
    &+\lambda \sum_{s=1}^{S}\lVert w^{(s)} \rVert_2^2+\alpha\sum_{s=1}^{S}\lVert w^{(s)} \rVert_1\\
    &+\beta \sum_{p=1}^{P}\lVert w^{(p)} \rVert_2^2.\\
    \label{Eqn: observed and potential outcome}
 \end{split}
\end{equation}

The first term in the expression above is the loss function for factual treatment assignment prediction. The second is the loss function for the prediction of observed outcomes. The third term is to balance representation distributions of treatment and control group. The fourth term is the elastic net term, used for deep feature selection and regularization. The last term regularizes the deep prediction network. By minimizing this objective function, FSRM obtains a nonlinear and balanced representation space, which can best predict individual treatment outcomes, mitigate selection bias, and ignore as much as possible the irrelevant variables. FSRM is implemented using standard feed-forward neural networks with Dropout ~\cite{srivastava2014dropout} and the ReLU activation function. Adam~\cite{kingma2014adam} is adopted to optimize the objective function. 

\subsubsection{Matching in Representation Space}
Leveraging the selective, balanced, and nonlinear representations extracted from observational data, we perform optimal matching~\cite{rosenbaum1989optimal} to find the matched samples with the smallest average absolute distance across all the matched pairs. In our work, the distance between treatment and control units is calculated based on Euclidean distance, Mahalanobis distance, and propensity score, respectively. The outcome of the selected control (treatment) unit within matched pair serves as the estimated counterfactual of the corresponding treatment (control) unit within each matched pair.

\section{Experiments}

In this section, we conduct experiments on three datasets, including the IHDP, modified IHDP, and a synthetic dataset, to evaluate the following aspects: (1) Our proposed method can improve treatment effect estimation with respect to average treatment effect and individual treatment effect. (2) The deep feature selection layers can help improve the performance of treatment effect estimation from observational data with high-dimensional variables or in the presence of different types of variables. (3) The proposed model is robust to different levels of treatment selection bias.

\subsection{Datasets}

\textbf{IHDP.} The IHDP dataset is a commonly adopted benchmark collected by the Infant Health and Development Program~\cite{brooks1992ihdp}. These data are generated based on a randomized controlled trial where intensive high-quality care and specialist home visits were provided to low-birthweight and premature infants. There are a total of 25 pre-treatment covariates and 747 units, including 608 control units and 139 treatment units. The outcome is the infants' cognitive test score which can be simulated using the pre-treatment covariates and the treatment assignment information through the NPCI package \footnote{\url{https://github.com/vdorie/npci}}. In the IHDP, a biased subset of the treatment group is removed to simulate the selection bias ~\cite{shalit2017estimating}. We repeat these procedures 1000 times to conduct evaluations of the uncertainty of estimates.

\noindent\textbf{Modified IHDP.} In practice, we cannot be sure that all collected variables are always relevant to the study. These variables can bring about extra uncertainty and bias. For IHDP, the outcome is simulated based on all the pre-treatment covariates in the IHDP dataset, so there are no irrelevant variables. Many observational studies include irrelevant variables that are related to neither the treatment nor the outcome of interest. To mimic this situation, we added 35 irrelevant variables as noise to increase data complexity. These additional variables, which are not associated with either the outcomes or the treatment assignments, are sampled from a multivariate normal distribution with mean 0 and random positive definite covariance matrix based on a uniform distribution over the space $35\times 35$ of the correlation matrix ~\cite{jacob2019group}. Compared to the original IHDP dataset, treatment effect estimation for the modified IHDP dataset is more challenging due to the irrelevant variables.  

\begin{figure}
    \centering
    \includegraphics[width=0.8\columnwidth]{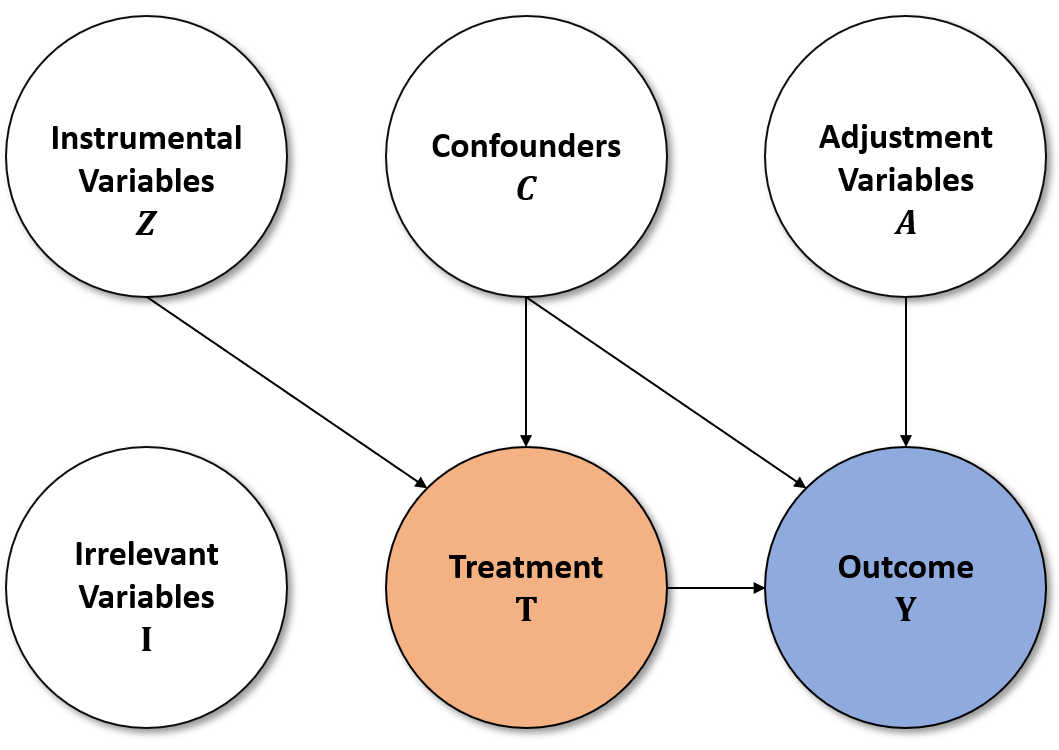}
    \caption{The types of observed variables.}
    \label{fig: decompose}
\end{figure}

\begin{table*}
  \caption{Performance comparison on IHDP, modified IHDP and synthetic data. We present the mean value $\pm$ the standard deviation of $\sqrt{\epsilon_\text{PEHE}}$ and $\epsilon_\text{ATE}$ on the test sets. We list available results reported by the original authors for IHDP dataset~\cite{schwab2018perfect,shalit2017estimating,yao2018representation}.}
  \label{IHDP}
  \centering
  \begin{tabular}{lllllll}
    \toprule
    \multicolumn{1}{c}{} & \multicolumn{2}{c}{IHDP} & \multicolumn{2}{c}{modified IHDP}  & \multicolumn{2}{c}{synthetic data}                  \\
    \cmidrule(lr){2-3} \cmidrule(lr){4-5} \cmidrule(lr){6-7}
    Method     & $\sqrt{\epsilon_\text{PEHE}}$   & $\epsilon_\text{ATE}$  & $\sqrt{\epsilon_\text{PEHE}}$     & $\epsilon_\text{ATE}$ & 
    $\sqrt{\epsilon_\text{PEHE}}$     & $\epsilon_\text{ATE}$\\
    \midrule
    kNN~\cite{ho2007matching} & $4.10 \pm 0.20$  & $0.79 \pm 0.05$   & $4.84 \pm 0.32$  & $0.84\pm 0.07$  & $0.40\pm0.04$ & $0.06\pm0.05$\\
    CF~\cite{wager2018estimation}     & $3.80\pm0.20$  & $0.40\pm0.03$   & $4.02\pm0.36$ & $0.47\pm0.06$  & $0.38\pm0.02$ & $0.05\pm0.03$ \\
    RF~\cite{breiman2001random}     & $6.60\pm0.30$  & $0.96\pm 0.06$   & $7.23\pm0.45$ & $1.02\pm0.07$  & $0.37\pm0.02$ & $0.04\pm0.03$\\
    BART~\cite{chipman2010bart}    & $2.30\pm0.10$  & $0.34\pm0.02$   & $3.65\pm0.21$ & $0.48\pm0.03$  & $0.38\pm0.02$ & $0.07\pm0.04$ \\
    GANITE~\cite{yoon2018ganite}     & $2.40\pm0.40$  & $0.49\pm 0.05$   & $3.71\pm0.43$ & $0.54\pm0.05$ & $0.48\pm0.05$ & $0.07\pm0.05$ \\
    PSM~\cite{ho2011matchit}     & $2.70\pm3.85$  & $0.49\pm0.81$   & $4.15\pm3.21$ & $1.35\pm0.75$   & $0.48\pm0.06$ & $0.20\pm0.13$\\
    TARNET~\cite{shalit2017estimating}     & $0.95\pm0.02$  & $0.28\pm0.01$   & $2.23\pm0.10$ & $0.48\pm0.03$  & $0.41\pm0.02$ & $0.06\pm0.04$ \\
    $\text{CFRNET}_\text{wass}$ ~\cite{shalit2017estimating} & $0.76\pm0.02$  & $0.27\pm0.01$   & $1.95\pm0.08$ & $0.38\pm0.02$  & $0.39\pm0.03$ & $0.06\pm0.04$   \\
    SITE~\cite{yao2018representation}  & $\textbf{0.66}\pm \textbf{0.11}$  & $0.20\pm0.01$   & $1.88\pm0.14$ & $0.36\pm0.02$  & $0.37\pm0.03$ & $0.05\pm0.04$ \\
    PM~\cite{schwab2018perfect}    & $0.84\pm0.61$  & $0.24\pm0.01$   & $2.12\pm0.53$ & $0.45\pm0.03$  & $0.37\pm0.04$ & $0.06\pm0.05$ \\
    \midrule
    FSRM$_\text{Mahal}$ & $1.32\pm0.05$  & $0.07\pm0.01$   & $1.39\pm0.08$ & $0.07\pm0.02$ & $0.15\pm0.01$ & $\textbf{0.01}\pm\textbf{0.01}$ \\
    FSRM$_\text{Euclid}$ & $1.24\pm0.04$  & $\textbf{0.05}\pm\textbf{0.01}$   & $\textbf{1.31}\pm\textbf{0.10}$ & $\textbf{0.06}\pm\textbf{0.01}$  & $\textbf{0.13}\pm\textbf{0.01}$ & $\textbf{0.01}\pm\textbf{0.01}$\\
    FSRM$_\text{Propensity}$ & $3.05\pm0.15$  & $0.09\pm0.01$   & $3.09\pm0.42$ & $0.09\pm0.02$& $0.20\pm0.01$ & $\textbf{0.01}\pm\textbf{0.01}$    \\
    \bottomrule
  \end{tabular}
\end{table*}

\noindent\textbf{Synthetic Dataset}.
To mimic situations where large numbers of variables and information on instrumental, adjustment, confounding, and irrelevant variables are available, we generate a synthetic dataset that reflects the complexity of observational medical records data. Our synthetic data includes confounders, instrumental, adjustment, and irrelevant variables. The interrelations among these variables, treatments, and outcomes are illustrated in Fig.~\ref{fig: decompose}. The number of observed variables in the vector $X= (C^\intercal,Z^\intercal,I^\intercal,A^\intercal)^\intercal$ is set to 60, including 15 confounders in $C$, 15 adjustment variables in $A$, 10 instrumental variables in $Z$, and 20 irrelevant variables in $I$. The model used to generate the continuous outcome variable $Y$ in this simulation is the partially linear regression model (Eq.~(\ref{Eqn: sim}))  extending the ideas described in ~\cite{jacob2019group, robinson1988root}: 

\begin{equation}
Y=\tau((C^\intercal, A^\intercal)^\intercal)T + g((C^\intercal, A^\intercal)^\intercal)+ \epsilon, \\
\label{Eqn: sim}
\end{equation}
where $T\overset{ind.}{\thicksim}\text{Bernoulli}(e_0((C^\intercal, Z^\intercal)^\intercal))$. $\epsilon$ are unobserved covariates, which follow a random standard normal distribution $N(0, 1)$ and $E[\epsilon|C,A,T]=0$.

We generate the confounders $C\in \mathbb{R}^{15}$, adjustment variables $A\in \mathbb{R}^{15}$, instrumental variables $Z\in \mathbb{R}^{10}$, and irrelevant variables $I\in \mathbb{R}^{20}$ in a way the variables in each type of variables are partially correlated among each other. They are generated by multivariate normal distribution with mean 0 and random positive definite covariance matrix based on a uniform distribution over the space $15\times15$, $15\times15$, $10\times10$, $20\times20$ of the correlation matrix, respectively ~\cite{jacob2019group}. The function $\tau((C^\intercal,A^\intercal)^\intercal)$ describes the true treatment effect as a function of the values of adjustment variables $A$ and confounders $C$; namely $\tau((C^\intercal, A^\intercal)^\intercal)=(\sin{((C^\intercal, A^\intercal)^\intercal \times b_{\tau})})^2$ where $b_{\tau}$ represents weights for every covariate in the function, which is generated by $\text{uniform}(0,1)$. The variable treatment effect implies that its strength differs among the units and is therefore conditioned on $C$ and $A$. The function $g((C^\intercal, A^\intercal)^\intercal)$ can have an influence on outcome regardless of treatment assignment. It is calculated via a trigonometric function to make the covariates non-linear, which is defined as $g((C^\intercal, A^\intercal)^\intercal)=(\cos{((C^\intercal, A^\intercal)^\intercal\times b_g)})^2$. Here, $ b_g$ represents a weight for each covariate in this function, which is generated by $\text{uniform}(0,1)$. The bias is attributed to unobserved covariates which follow a random normal distribution $N(0, 1)$. The treatment assignment $T$ follows the Bernoulli distribution, i.e., $T\overset{ind.}{\thicksim}\text{Bernoulli}(e_0((C^\intercal, Z^\intercal)^\intercal))$  with probability $e_0((C^\intercal, Z^\intercal)^\intercal) = \Phi(\frac{a-\mu(a)}{\sigma(a)})$, where $e_0((C^\intercal, Z^\intercal)^\intercal)$ represents the propensity score, which is the cumulative distribution function for a standard normal random variable based on confounders $C$ and instrumental variables $Z$, i.e., $a = \sin{((C^\intercal, Z^\intercal)^\intercal\times b_a)}$, where $b_a$ is generated by $\text{uniform}(0,1)$.

The total sample size in our synthetic data is 2000, including 1000 units in the treatment group and 1000 units in the control group. During our simulation procedure, $e_0((C^\intercal, Z^\intercal)^\intercal)$ is the propensity score, which represents the treatment selection bias based on their own confounders $C$ and instrumental variables $Z$. We randomly draw 750 units in the control group and 250 in the treatment group to compose a synthetic dataset with 1000 units.  To ensure a robust estimation of model performance, we repeat the random sampling procedure 1000 times and obtain 1000 synthetic datasets.

\subsection{Baseline Methods}

We compare the proposed feature selection representation matching (FSRM) method with the following baseline methods. 

\textbf{k-nearest neighbor (kNN)} method performs matching in the covariate space and uses nonparametric preprocessing matching to reduce model dependence in parametric causal inference~\cite{ho2007matching}.
 \textbf{Causal forests (CF)} is a nonparametric forest-based method for estimating heterogeneous treatment effects by extending Breiman's random forest algorithm~\cite{wager2018estimation}.
 \textbf{Random forest (RF)} is a classifier consisting of a combination of tree predictors, in which each tree depends on a random vector that is independently sampled and has the identical distribution for all trees~\cite{breiman2001random}.
\textbf{Bayesian additive regression trees (BART)} is a nonparametric Bayesian regression model, which uses dimensionally adaptive random basis elements. Every tree in BART model is a weak learner, and it is constrained by a regularization prior. Information can be extracted from the posterior by a Bayesian backfitting MCMC algorithm~\cite{chipman2010bart}.
 \textbf{Generative adversarial nets for inference of ITE (GANITE)} is based on the Generative Adversarial Nets framework. It generates proxies of the counterfactual outcomes using a counterfactual generator and then passes these proxies to an ITE generator~\cite{yoon2018ganite}.
 \textbf{Propensity score matching (PSM)} conducts a matching based on a predicted probability of group membership, which is obtained from logistic regression based on covariates~\cite{ho2011matchit}.
 \textbf{Treatment-agnostic representation network (TARNET)} is a variant of counterfactual regression without the balance regularization ~\cite{shalit2017estimating}.
  \textbf{Counterfactual regression ($\text{CFRNET}_\text{wass}$)} maps the original features into a latent representation space by minimizing the error in predicting factual outcomes and imbalance measured by Wasserstein distance between the treatment representations and the control representations~\cite{shalit2017estimating}.
 \textbf{Local similarity preserved individual treatment effect estimation method (SITE)} is a deep representation learning, which preserves local similarity and balances data distributions simultaneously, by focusing on several hard samples in each mini-batch~\cite{yao2018representation}.
 \textbf{Perfect match (PM)} augments samples within a minibatch with their propensity-matched nearest neighbors and then implements existing neural network architectures~\cite{schwab2018perfect}.

By using different distance metrics in matching, the proposed FSRM method has three variants denoted as FSRM$_\text{Mahal}$, FSRM$_\text{Euclid}$ and FSRM$_\text{Propensity}$, which adopt the Mahalanobis distance, Euclidean distance, and propensity score, respectively.

\subsection{Parameter Settings}

The parameters of baseline methods are set the same as suggested in the original papers. To ensure a fair comparison, we follow a standardized approach ~\cite{schwab2018perfect} to hyperparameter optimization for modified IHDP and the synthetic datasets. The hyperparameters of our method are chosen based on performance on the validation dataset, and the searching range is shown in Table~\ref{hyperparameter}.

\begin{table}[th!]
  \caption{Hyperparameters and ranges.}
  \label{hyperparameter}
  \centering
  \begin{tabular}{ll}
    \toprule
    \multicolumn{1}{c}{Hyperparameter} & \multicolumn{1}{c}{Range}            \\
    \midrule
    $\delta$   &  0, $\{10^{k}\}_{k=-6}^{2}$, 0.2 , 0.5, 2, 5 \\ 
    \midrule
    $\gamma$, $\lambda$,  $\alpha$,  $\beta$ & 0, $\{10^{k}\}_{k=-6}^{0}$, 0.2, 0.5\\
    \midrule
     & (dim. of input, 200, 150, 100)\\
    No. and dim. of deep & (dim. of input, 200, 100, 50)\\
    feature selection layers & (dim. of input, 100, 100)\\
     & (dim. of input, 100, 50)\\
    \midrule
    No. of deep prediction layers & 1, 2, 3, 4\\
    \midrule
    Dim. of deep prediction layer & 50, 100, 150\\
    \midrule
    Batch size & 100, 200, 300\\
    \bottomrule
  \end{tabular}
\end{table}

\subsection{Results and Analysis}
For the IHDP, modified IHDP, and our synthetic datasets, we adopt two commonly used evaluation metrics. The first one is the error of ATE estimation, which is defined as:
\begin{equation}
    \epsilon_\text{ATE}  = |\text{ATE} - \widehat{\text{ATE}}|,
\end{equation}
where \text{ATE} is the true value and $\widehat{\text{ATE}}$ is an estimated \text{ATE}. 

The second one is the error of expected precision in estimation of heterogeneous effect (PEHE)~\cite{hill2011bayesian}, which is defined as:
\begin{equation}
    \epsilon_\text{PEHE}  = \frac{1}{n}\sum_{i=1}^{n}(\text{ITE}_i-\widehat{\text{ITE}}_i)^2,
\end{equation}
where $\text{ITE}_i$ is the true \text{ITE} for unit $i$ and $\widehat{\text{ITE}}_i$ is an estimated \text{ITE} for unit $i$.

Table~\ref{IHDP} shows the performance of our method and baseline methods on the IHDP, the modified IHDP, and our synthetic datasets over 1000 realizations. We report the average results and also the standard deviations. FSRM with the Euclidean distance achieves the best performance with respect to $\epsilon_\text{ATE}$ for all three datasets and the best performance with respect to $\sqrt{\epsilon_\text{PEHE}}$ in the modified IHDP and our synthetic dataset. For $\sqrt{\epsilon_\text{PEHE}}$ in IHDP, FSRM is better than kNN, CF, RF, BART, GANITE, and PSM, but is outperformed by TARNET, CFRNET, SITE, and PM. Compared with regression-based methods, FSRM is not the top performer, but this result is already remarkable in matching-based methods with respect to the individual treatment effect estimation. In addition, because in the original IHDP data, all of the variables are treated as pre-treatment and there are no irrelevant variables, it cannot fully demonstrate the advantages of our method. The inclusion of the feature selection layer in FSRM is not expected to come without cost when all variables are relevant as in the case of IHDP which has no irrelevant variables. This is backed up by the significant gains of FSRM in the modified IHDP dataset in which irrelevant variables are added to IHDP lending extra complexity to the data. Our method can effectively filter out these noises and remain fairly steady in estimating treatment effect with respect to both $\sqrt{\epsilon_\text{PEHE}}$ and $\epsilon_\text{ATE}$ for the IHDP and modified IHDP. However, the other methods suffered a marked decline in performance for the modified IHDP dataset. Moreover, because our synthetic dataset includes several different types of variables such as instrumental variables, adjustment variables, confounders, and irrelevant variables, our model demonstrates clear superiority over the state-of-the-art methods when dealing with a large number of variables with complicated relationships among them. In addition, we find that $\text{FSRM}_\text{Euclid}$ performs better than $\text{FSRM}_\text{Mahal}$ and $\text{FSRM}_\text{Propensity}$ for all three datasets. Euclidean distance better suits optimal matching based on representation vectors.

\subsection{Model Evaluation}
\label{Model Evaluation}
Experimental results on three datasets show that FSRM provides more accurate estimations of average treatment effect and individual treatment effect, and it is more highly adaptable to complicated observational data than the state-of-the-art matching estimators and representation learning methods. We further evaluate the performance of FSRM from three perspectives including robustness with respect to different levels of treatment selection bias, the effectiveness of each component of the proposed FSRM, and feature selection interpretability.

\begin{figure}
    \centering
    \includegraphics[width=0.85\columnwidth]{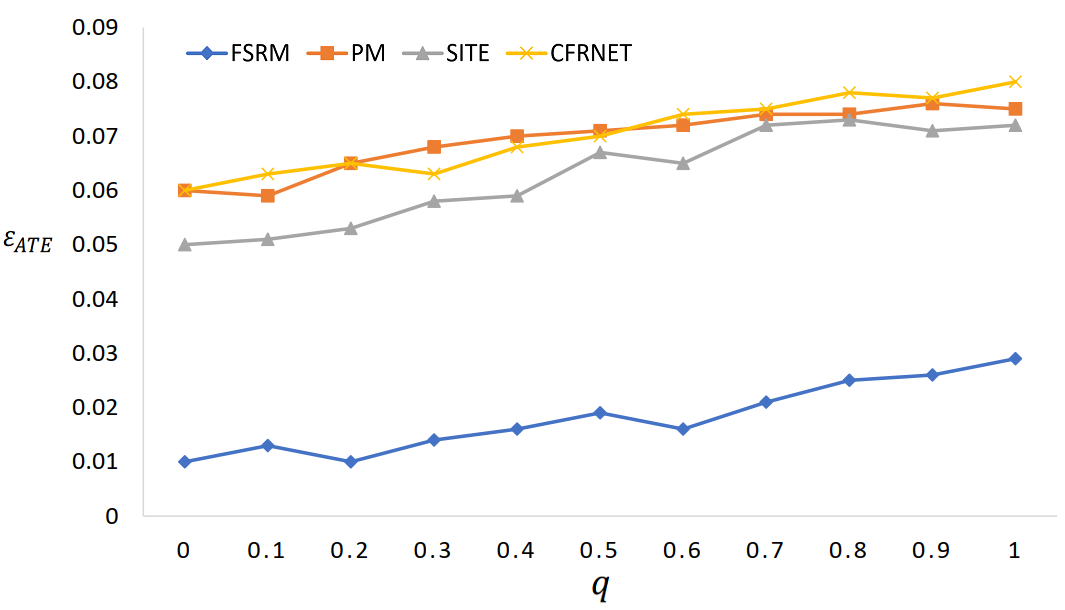}
    \caption{$\sqrt{\epsilon_\text{PEHE}}$ and $\epsilon_\text{ATE} $ performance on simulation dataset with $q$ from 0 to 1.}
    \label{fig: biasate}
\end{figure}

Firstly, we evaluate the robustness of our proposed method with respect to different levels of treatment selection bias. Although in our simulation procedure, the treatment selection bias has been taken into account based on their own propensity score $e_0((C^\intercal, Z^\intercal)^\intercal)$, we use conditional sampling from treatment and control groups to increase the treatment selection bias. If the propensity score $e_0$ is equal to constant 0.5, it means no matter what the confounders and instrumental variables are, the unit is randomly assigned to either the treatment or the control group with the same probability, so that there is no treatment selection bias. The greater $|e_0((C^\intercal, Z^\intercal)^\intercal)-0.5|$ is, the larger selection bias will end up getting. Following the idea in ~\cite{shalit2017estimating}, with probability $1-q$, we randomly draw the treatment and control units; with probability $q$, we draw the treatment and control units that have the greatest $|e_0((C^\intercal, Z^\intercal)^\intercal)-0.5|$. Thus, the higher the $q$ is, the larger the selection bias is. We run $\text{CFRNET}_\text{WASS}$, SITE, PM and our method $\text{FSRM}_\text{Euclid}$ on the simulation datasets with $q$ from 0 to 1, and show the results in Fig.~\ref{fig: biasate}. We can observe that our method consistently outperforms the baseline methods under different levels of divergence and is robust to a high level of treatment assignment bias. 

\begin{figure}
    \centering
    \includegraphics[width=0.85\columnwidth]{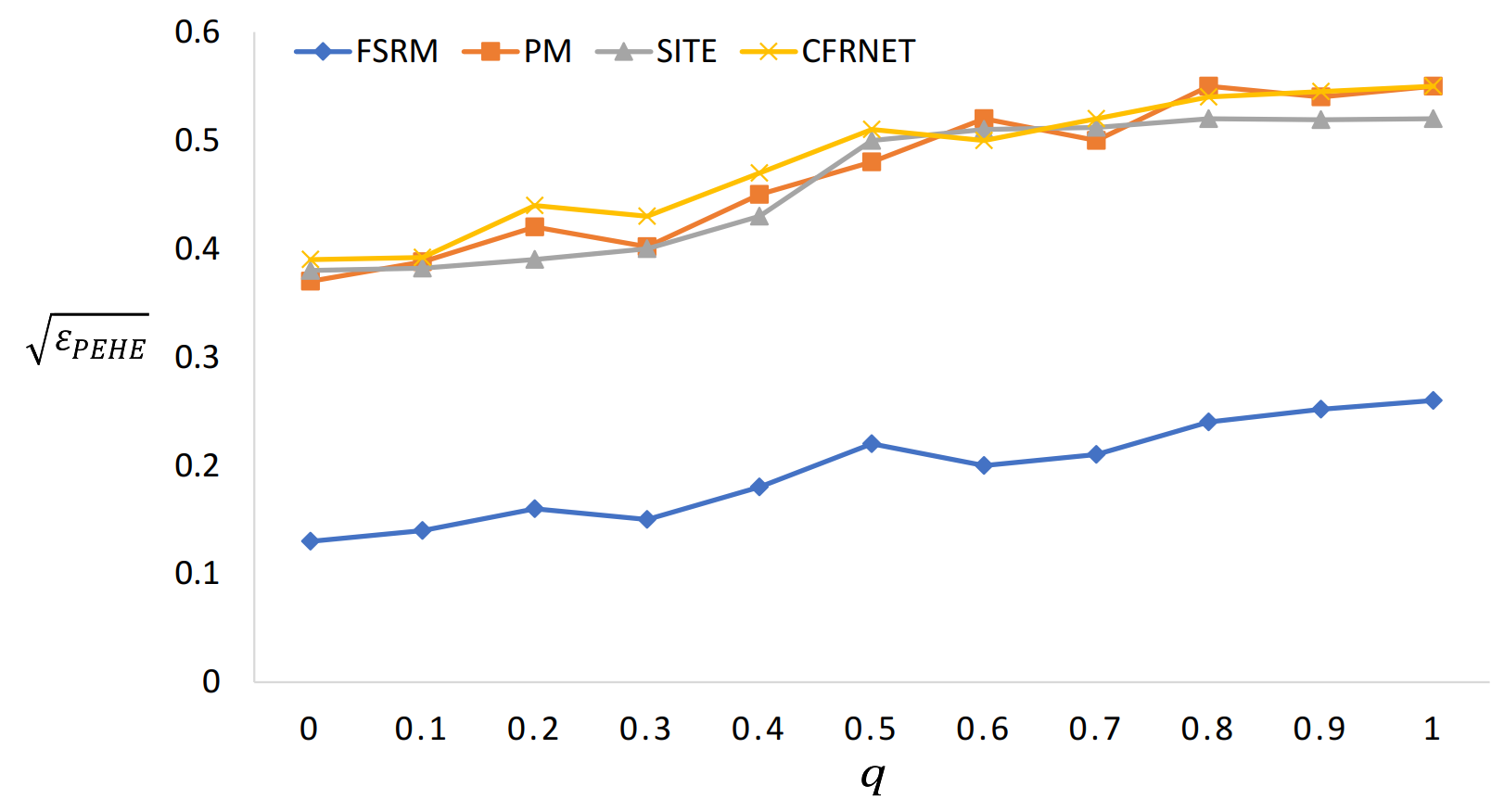}
    \label{fig: biaspehe}
\end{figure}

In addition, we trained two ablation studies of FSRM$_\text{Euclid}$ on our synthetic dataset. The first one is FSRM (w/o FSL) where the sparse one-to-one feature selection layer between the input and the first hidden layer, and elastic net throughout the fully connected representation layers are removed. We only use a normal fully connected neural network to learn the representation space. The second ablation study is FSRM (w/o IPM) where the integral probability metric is removed and there is not any restriction on the divergence between the representation distributions of treatment and control groups. 

As shown in Table~\ref{ablation}, the performance becomes poor after removing either the feature selection layers or the IPM module compared to the original FSRM. More specifically, after removing the feature selection layers,  $\sqrt{\epsilon_\text{PEHE}}$ and $\epsilon_\text{ATE}$ increase dramatically and have similar performance to other baseline methods. Working with the original synthetic data where treatment selection bias is from own propensity score, i.e., $T\overset{ind.}{\thicksim}\text{Bernoulli}(e_0((C^\intercal, Z^\intercal)^\intercal))$, removing the IPM module from the FSRM (w/o IPM) only has a little impact on $\sqrt{\epsilon_\text{PEHE}}$ and $\epsilon_\text{ATE}$.  However, if extra bias (q=0.5 and q=1) is added to the synthetic data, the FSRM (w/o IPM) has poor performance compared to the original FSRM. In addition, as the extra bias increases, the difference between the performance of FSRM (w/o IPM) and the original FSRM increases further. Therefore, the feature selection layers and IPM module are essential components of our model.

\begin{table}[t]
  \caption{Summary of results in ablation studies.}
  \label{ablation}
  \centering
  \begin{tabular}{llll}
    \toprule
    \multicolumn{1}{c}{Synthetic data} & \multicolumn{1}{c}{Method} & \multicolumn{1}{c}{$\sqrt{\epsilon_\text{PEHE}}$} & \multicolumn{1}{c}{$\epsilon_\text{ATE}$}   \\
    \midrule
    Original bias  & FSRM & $0.13\pm0.01$  & $0.01\pm 0.01$   \\
    (q=0) & FSRM (w/o FSL) & $0.41\pm0.02$  & $0.04\pm0.01$   \\
    & FSRM (w/o IPM) & $0.15\pm0.01$  & $0.01\pm0.01$  \\
    \midrule
    Extra bias  & FSRM & $0.22\pm0.01$  & $0.02\pm0.01$    \\
    (q=0.5)& FSRM (w/o IPM) & $0.34\pm0.04$  & $0.03\pm0.01$   \\
    \midrule
    Extra bias & FSRM  & $0.26\pm0.01$  & $0.03\pm0.01$ \\
    (q=1) & FSRM (w/o IPM) & $0.58\pm0.04$  & $0.07\pm0.03$  \\
    \bottomrule
  \end{tabular}
\end{table}

\begin{figure}[th!]
    \centering
    \includegraphics[width=1\columnwidth]{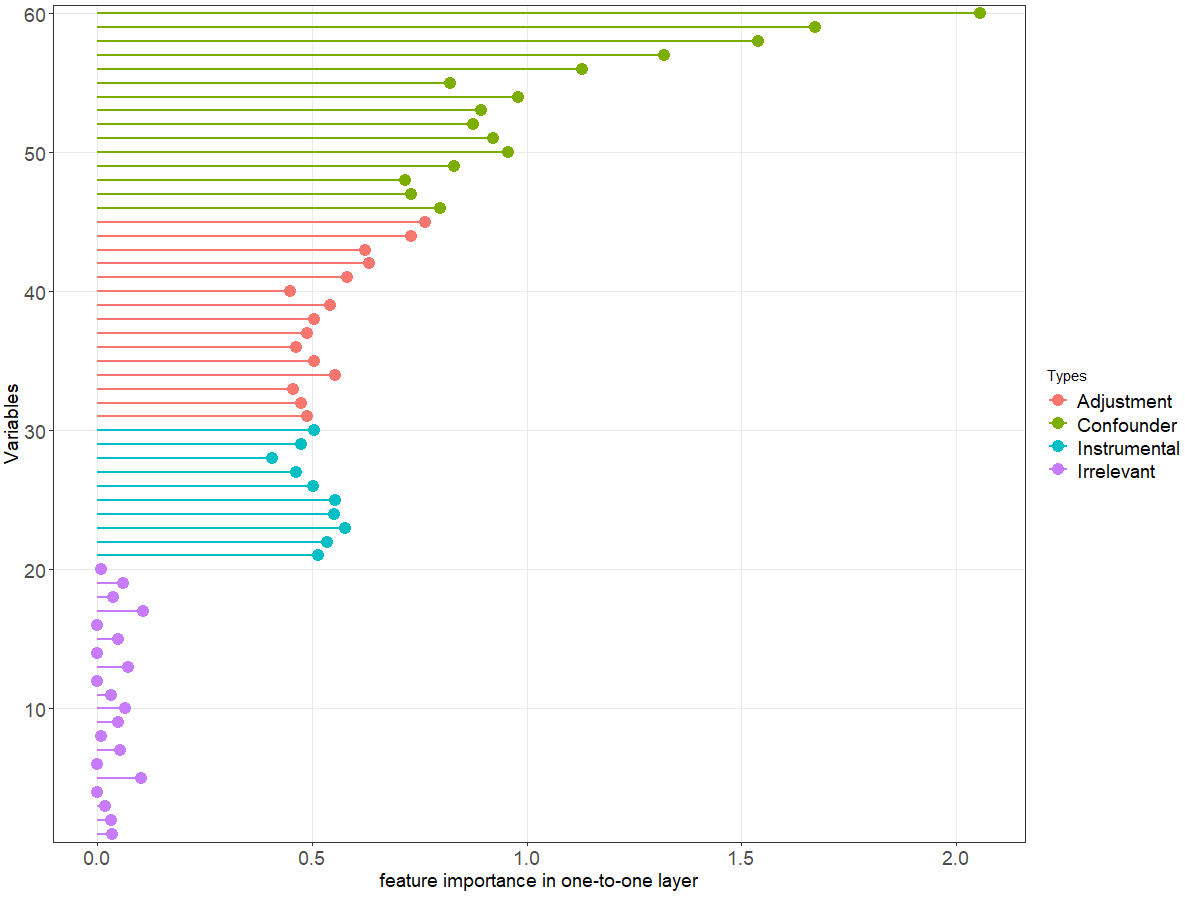}
    \caption{Feature importance in one-to-one feature selection layer for 60 input variables.}
    \label{fig: importance}
\end{figure}

For feature selection, we have two parts: a sparse one-to-one feature selection layer between the input and the first hidden layer, and the elastic net throughout the fully connected representation layers. That one-to-one feature selection layer at the input level selects which variables are input into the neural network, which makes the deep neutral network more interpretable. In the one-to-one feature selection layer instead of fully connected layers, every input variable only connects to its corresponding node where the input variable is weighted. This weight can give us an intuitive impression of feature importance for each variable.  Fig.~\ref{fig: importance} shows the importance of each feature in the one-to-one layer for the 60 input variables. As we expect, smaller or even zero weights are assigned to irrelevant variables, and the largest weights are assigned to the confounders. Although the elastic net will continue to conduct feature selection, this figure can give us one clear impression of which variables are forwarded into the following ``black box'' deeper layers of the deep neural network.

\subsection{Sensitivity Analysis}

\begin{figure}[t]
  \centering
  \begin{minipage}[b]{0.38\textwidth}
    \includegraphics[width=\textwidth]{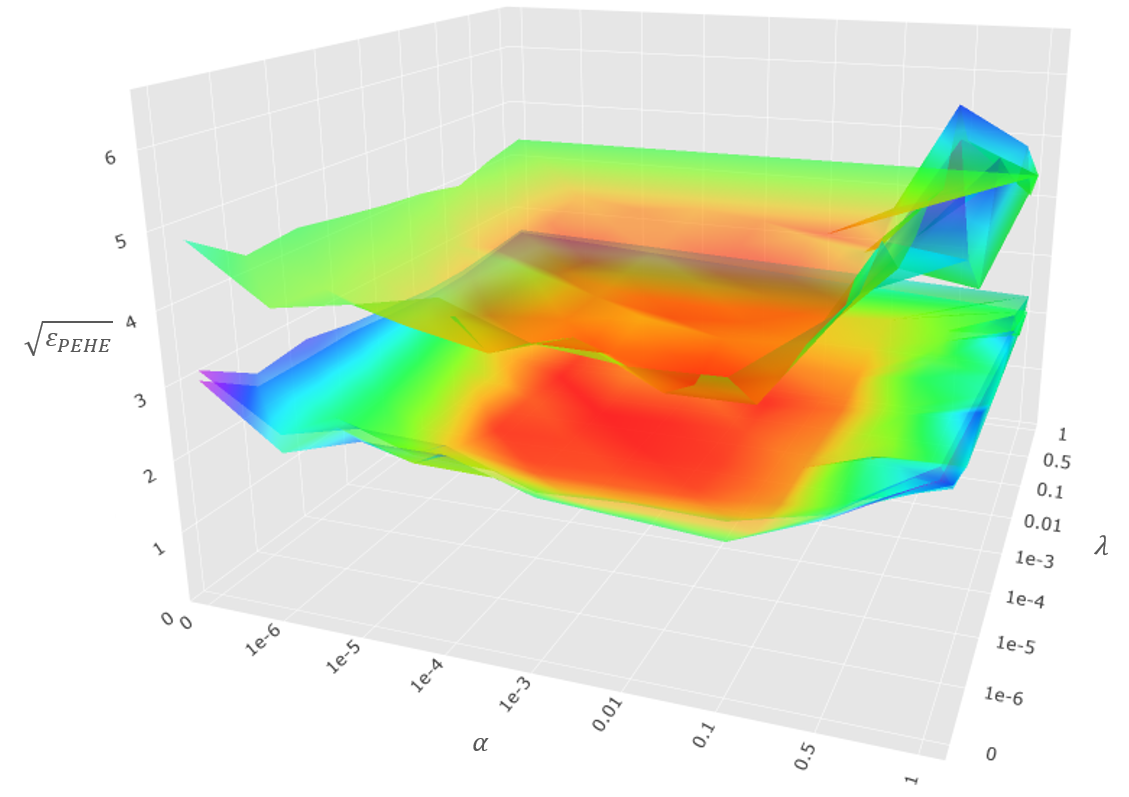}
    %\caption{ performance on combinations of hyper-parameters of $L_1$ and $L_2$ penalty. There are three layers, which represent $\text{FSRM}_\text{Propensity}$,  $\text{FSRM}_\text{Mahal}$ and $\text{FSRM}_\text{Euclid}$ from top to bottom.}
    \label{fig: l1l2pehe}
  \end{minipage}
  \hfill
  \begin{minipage}[b]{0.38\textwidth}
    \includegraphics[width=\textwidth]{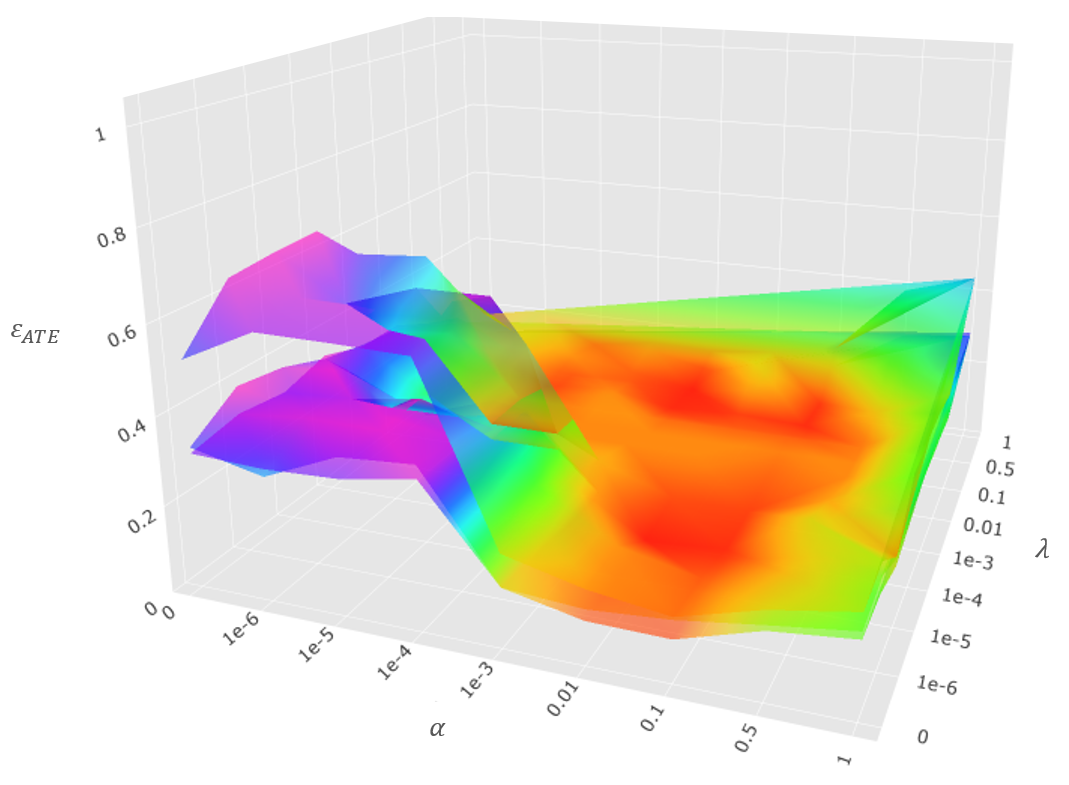}
    \caption{$\sqrt{\epsilon_\text{PEHE}}$ and $\epsilon_\text{ATE} $ performance on combinations of hyper-parameters of $L_1$ and $L_2$ penalty. There are three layers, which represent $\text{FSRM}_\text{Propensity}$,  $\text{FSRM}_\text{Mahal}$ and $\text{FSRM}_\text{Euclid}$ from top to bottom.}
    \label{fig: l1l2ate}
  \end{minipage}
\end{figure}

\begin{figure}[t]
  \centering
  \begin{minipage}[b]{0.4\textwidth}
    \includegraphics[width=\textwidth]{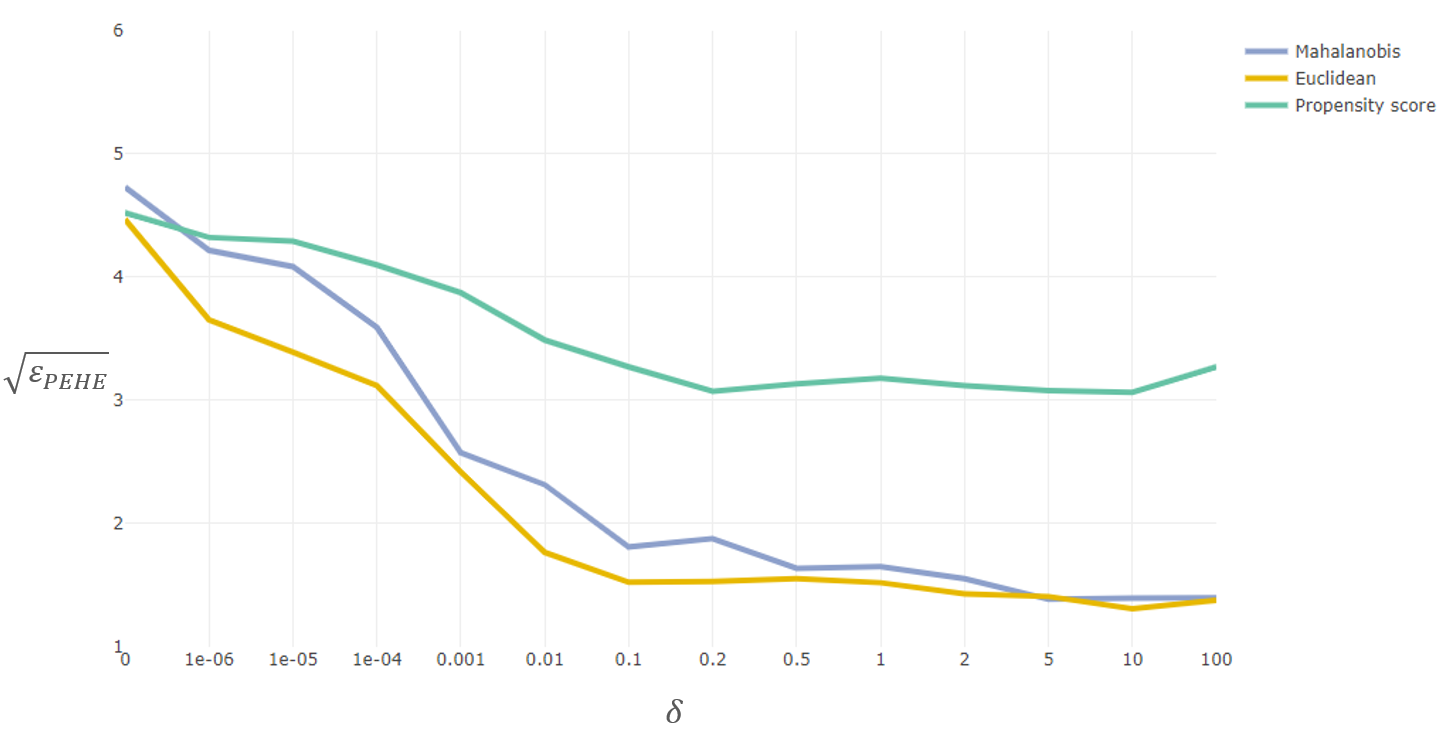}
    %\caption{$\sqrt{\epsilon_\text{PEHE}}$ performance on hyper-parameters $\delta$.}
    \label{fig: ypehe}
  \end{minipage}
  \hfill
  \begin{minipage}[b]{0.4\textwidth}
    \includegraphics[width=\textwidth]{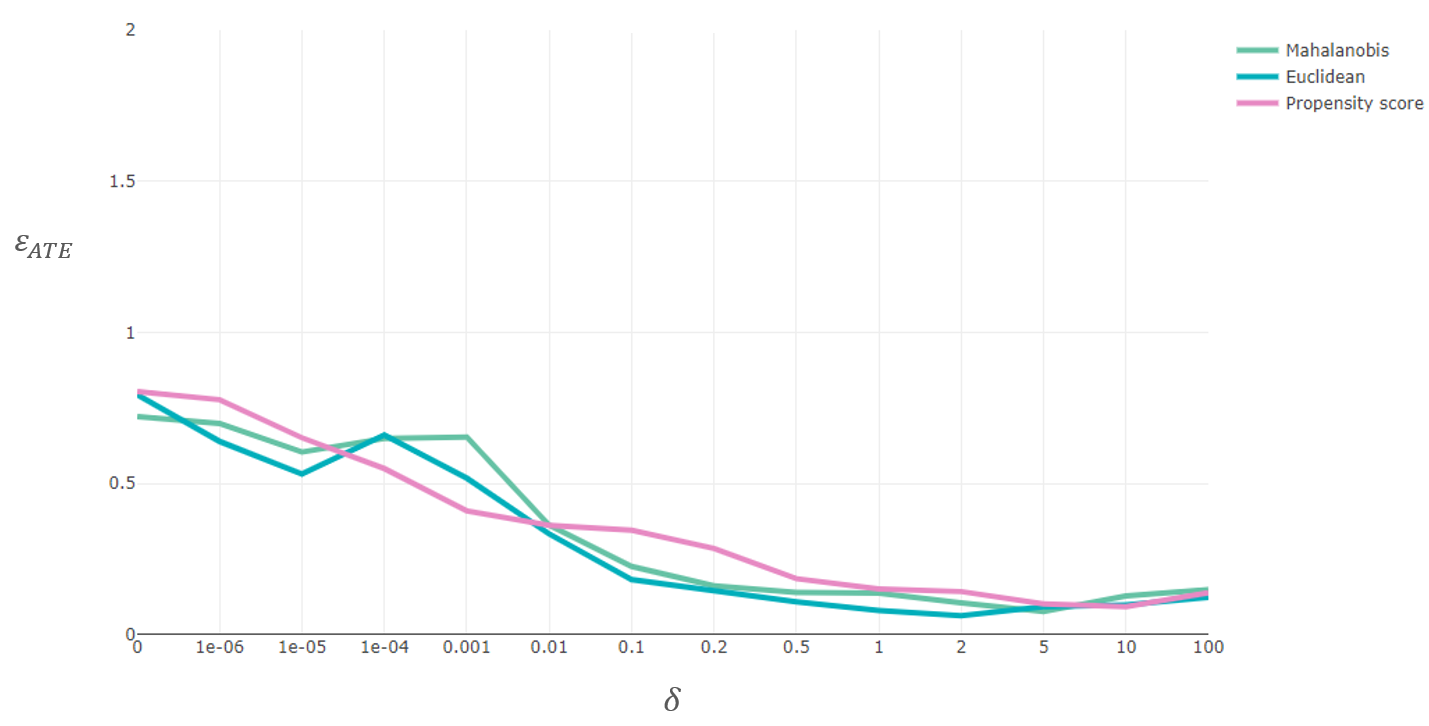}
    \caption{$\sqrt{\epsilon_\text{PEHE}}$ and $\epsilon_\text{ATE} $ performance on hyper-parameters $\delta$.}
    \label{fig: yate}
  \end{minipage}
\end{figure}

We evaluate FSRM's sensitivity to the three most important parameters $\delta, \lambda\ \text{and} \ \alpha $ on modified IHDP dataset, which respectively control the weights of observed outcome prediction, smoothness, and sparsity of the weights in the feature selection layer. Based on our analysis presented in Fig.~\ref{fig: l1l2ate}, the performance of our model, in terms of $\sqrt{\epsilon_\text{PEHE}}$ and $\epsilon_\text{ATE}$, is significantly improved compared with the model without $L_1$ and $L_2$ penalties. Also, the overall performance on different combinations of hyperparameters of $L_1$ and $L_2$ penalties is stable over a large parameter range, which confirms the effectiveness and robustness of deep feature selection in FSRM. This conclusion is consistent with our model evaluation results.

In Fig.~\ref{fig: yate}, we find that adding prediction of observed outcome into the model can significantly improve the performance in terms of $\sqrt{\epsilon_\text{PEHE}}$ and $\epsilon_\text{ATE}$, compared with only having the prediction of treatment assignment (i.e., $\delta = 0$). This is the main reason why our method performs well when estimating individual treatment effects and average treatment effects, but traditional propensity score matching with logistic regression predicting the treatment assignment cannot accurately estimate individual treatment effects.

\section{Related Work}

Embracing the rapid developments in machine learning and deep learning, various causal effect estimation methods for observational data have sprung up.  Balancing neural networks (BNNs) ~\cite{johansson2016learning} and counterfactual regression networks (CFRNET) ~\cite{shalit2017estimating} are proposed to balance covariate distributions across treatment and control groups by formulating the problem of counterfactual inference as a domain adaptation problem. This model is extended to any number of treatments even with continuous parameters, as described in the perfect match (PM) approach~\cite{schwab2018perfect} and DRNets ~\cite{schwab2019learning}. Following this idea, a few improved models have been proposed and discussed. For example, the shift-invariant representation learning is combined with re-weighting methods ~\cite{johansson2018weighted_rep}. A local similarity preserved individualized treatment effect (SITE) estimation method ~\cite{yao2018representation} is proposed focusing on local similarity information that provides meaningful constraints on individual treatment estimation. Generative adversarial networks ~\cite{yoon2018ganite} for individualized treatment effects have also been proposed for individual treatment effect estimation. 

Besides the popular representation learning methods, matching methods are also among the most widely used approaches to causal inference from observational data. The core purpose of matching methods is to reduce the estimation bias brought on by confounders, so how to find the most similar neighbors in the opposite treatment group is the most important problem. The similarity among neighbors can be measured by different distance metrics, including distance based on original covariates space such as the Euclidean distance~\cite{rubin1973matching} and Mahalanobis distance ~\cite{rubin2000combining}, and distance based on transformed space such as the propensity score~\cite{rosenbaum1983central}, prognosis score ~\cite{hansen2008prognostic}, random subspaces~\cite{li2016matching_digital_ijcai16}, and balanced and nonlinear representation based nearest neighbor matching (BNR-NNM)~\cite{li2017matching_nips17}.

We propose a feature selection representation matching method in this paper, which inherits the advantages of both the matching based methods and the representation learning based methods. Different from existing work on treatment effect estimation, our method learns a selective, nonlinear, and balanced representation through deep neural networks, and performs matching in the representation space. Incorporating feature selection layers into deep representation learning, which simultaneously predicts the treatment assignment and outcomes, makes the representation space best predictive of individual treatment outcome, mitigates treatment selection bias, and minimizes the influence of irrelevant variables.

\section{Conclusions}

In this paper, we present a novel feature selection representation matching (FSRM) method for estimating individual treatment effect and average treatment effect, which combines the predictive power of deep learning and interpretability of matching methods. It is applicable to and has good performance for observational data especially with high-dimensional variables or in the presence of different types of variables. Experimental results on three datasets show that FSRM provides more accurate estimations of average treatment effect and individual treatment effect, and is more highly adaptable to complicated observational data than the state-of-the-art matching estimators and representation learning methods. 

\balance
\bibliographystyle{ACM-Reference-Format}
%%% -*-BibTeX-*-
%%% Do NOT edit. File created by BibTeX with style
%%% ACM-Reference-Format-Journals [18-Jan-2012].

%%
%% If your work has an appendix, this is the place to put it.

\end{document}